\documentclass{article}
\usepackage{iclr2026_conference,times}

\usepackage{amsmath,amsfonts,bm}

\def\eqref#1{equation~\ref{#1}}

\def\1{\bm{1}}

\DeclareMathAlphabet{\mathsfit}{\encodingdefault}{\sfdefault}{m}{sl}
\SetMathAlphabet{\mathsfit}{bold}{\encodingdefault}{\sfdefault}{bx}{n}

\usepackage{amsmath}
\usepackage{amssymb}
\usepackage{booktabs}
\usepackage{graphicx}
\usepackage{flafter}
\usepackage[section]{placeins}
\usepackage{makecell}
\usepackage{multirow}
\usepackage{tabularx}
\usepackage{xcolor}
\usepackage{url}
\usepackage{hyperref}
\usepackage{booktabs}

\usepackage{array}
\usepackage{rotating}
\usepackage{tabularx}

\hypersetup{
colorlinks=true,
linkcolor=red,
citecolor=cyan,
filecolor=magenta,      
urlcolor=magenta,
}

\title{Is Multimodal Speculative Decoding Ready for Diffusion-Based Parallel Drafting? A Survey and Empirical Diagnosis}

\author{\textbf{Yantao Li}$^{1,2,3,*}$,
\textbf{Huanlin Gao}$^{2,3,*}$,
\textbf{Fang Zhao}$^{2,3,*}$,
\textbf{Chao Tan}$^{2,3}$,
\textbf{Qiang Hui}$^{2,3}$,
\textbf{Shuting Liu}$^{2,3}$,\\
\textbf{Fuyuan Shi}$^{2,3}$,
\textbf{Ting Lu}$^{2,3}$,
\textbf{Shaoan Zhao}$^{2,3}$,
\textbf{Xueqiang Guo}$^{2,3}$,
\textbf{Xinpei Su}$^{2,3}$,
\textbf{Jianbing Zhang}$^{1}$,\\
\textbf{Xinyu Dai}$^{1,\dagger}$,
\textbf{Kai Wang}$^{2,3,\ddagger}$,
\textbf{Shiguo Lian}$^{2,3,\dagger}$
\\
\textbf{ }\\[-0.1cm]
$^{1}$National Key Laboratory for Novel Software Technology, Nanjing University\\
$^{2}$Data Science \& Artificial Intelligence Research Institute, China Unicom\\
$^{3}$Unicom Data Intelligence, China Unicom\\
}

\iclrfinalcopy

\begin{document}

\renewcommand{\thefootnote}{}
\footnotetext[0]{%
\textsuperscript{*}Equal contribution.
\textsuperscript{$\dagger$}Corresponding author.
\textsuperscript{$\ddagger$}Project leader.
}
\maketitle
\begin{abstract}

Speculative decoding accelerates autoregressive generation by allowing a lightweight drafter to propose future tokens while a target model verifies them in parallel. Its lossless guarantee has motivated a line of work that pushes the drafter itself toward parallel generation—the most recent prominent milestone is block-parallel generative approaches, including diffusion-based drafting methods such as DFlash and DSpark, achieving a 3.6$\times$ speedup on common daily conversation tasks.
While diffusion-based block-parallel drafting has been explored in text-only LLMs, its transferability to multimodal models remains an open question.
Existing multimodal speculative decoding efforts focus on input compression, adapter alignment, candidate coverage, or modality-specific verification; however, block-parallel generative drafting remains largely unexplored.
To bridge this gap, this paper combines a modality-centered survey with a cross-architecture empirical study to ask: Is multimodal speculative decoding ready for diffusion-based parallel drafting? In this survey, we systematically analyze a wide spectrum of multimodal models—spanning Vision-Language, Video-Language, Audio, and Vision-Language-Action (VLA) architectures—from the dual perspectives of drafting parallelism and cross-modal information interaction. We introduce a unified taxonomy that isolates drafter-side parallelism from orthogonal design choices such as tree construction and verification strategies. Furthermore, we provide a comprehensive empirical evaluation of existing methods under varying degrees of parallelism across standardized multimodal benchmarks, such as OCR, VQA, visual reasoning, and image captioning. Finally, we summarize the limitations of current approaches, discuss open challenges, and outline promising future directions for this rapidly evolving field.

\end{abstract}
\section{Introduction}
\label{sec:introduction}

Multimodal generative models have achieved remarkable progress in a wide
range of applications, including visual question answering, document
understanding, chart reasoning, video analysis, speech interaction, and
embodied control. Despite these advances, inference remains dominated by
autoregressive decoding. After multimodal context encoding and prefill, the
decoder still generates tokens sequentially, requiring one expensive target
model execution for each output token. This limitation becomes increasingly
severe for long-form reasoning, OCR-heavy generation, structured documents,
and interactive multimodal agents.

Speculative decoding reduces this autoregressive bottleneck by introducing a
lightweight drafter that proposes future tokens, which are then verified by
the target model in parallel~\citep{leviathan2023fast,chen2023accelerating}.
The achievable acceleration depends on two complementary factors: how many
draft tokens can be accepted by the target model and how efficiently the draft candidates can be generated. 
Existing speculative decoding methods have therefore evolved along two directions.

The first direction improves verification efficiency by increasing candidate
coverage. Tree-based speculative decoding methods, including
SpecInfer~\citep{miao2024specinfer}, Sequoia~\citep{chen2024sequoia},
DDTree~\citep{ringel2026ddtree}, and TAPS~\citep{wang2026taps}, organize
multiple candidate continuations into tree structures and verify them in
parallel, improving the expected accepted tokens under a fixed verification
budget. These approaches mainly optimize the utilization of each target model
forward pass while retaining sequential or weakly parallel drafting.

The second direction improves drafter-side parallelism. Traditional
autoregressive drafters generate a block of $K$ future tokens through $K$
serial draft steps. Feature-based approaches such as EAGLE
\citep{li2024eagle1,li2024eagle2,li2025eagle3}, multi-head prediction methods
such as Medusa~\citep{cai2024medusa}, and multi-token prediction
\citep{gloeckle2024better} reduce drafting latency by predicting multiple
future positions in parallel. More recently, diffusion-based block-parallel
drafting has emerged as a stronger form of output-side parallelism. DFlash
\citep{chen2026dflash} formulates future-token generation as a block-level
diffusion process, while DSpark~\citep{cheng2026dspark} introduces lightweight
intra-block dependency modeling and confidence-aware scheduling. These
methods shift speculative decoding from token-wise prediction toward jointly
generating future blocks.

Together, these advances change the central question of speculative decoding
from ``Can the target verify multiple tokens simultaneously?'' to ``Can the
drafter generate sufficiently accurate future blocks without paying the cost
of serial drafting?''

Multimodal speculative decoding introduces additional challenges beyond
text-only generation. Existing approaches mainly investigate how a drafter
should access visual information, how redundant visual tokens should be
compressed, how target representations can improve draft quality, and how
verification should account for perceptual or functional equivalence
\citep{gagrani2024multimodal,ganesan2025massv,hu2025dream,huang2025specvlm,
wang2025specvla}. These studies demonstrate that multimodal conditioning is
critical for maintaining draft--target agreement. However, most existing
methods remain based on token-wise or semi-parallel drafting. Whether
block-parallel drafting can provide practical acceleration for frozen,
off-the-shelf multimodal targets under exact verification remains largely
unexplored.

To analyze this question systematically, we introduce a three-level taxonomy
based on drafter-side parallelism. L0 advances one future token per forward
step, covering autoregressive drafters and feature-based approaches such as
EAGLE. L1 predicts multiple predefined future positions from a shared prefix,
including methods such as Medusa and multi-token prediction. L2 treats the
future block as a jointly generated or refined unit, including recent
block-parallel methods such as DFlash and DSpark. Candidate organization
strategies, such as tree branching and verification rules, are orthogonal to
this taxonomy. Under this view, most existing multimodal speculative decoding
methods operate at L0 or L1. Existing L2 multimodal systems either modify the
target architecture for self-speculation
\citep{wu2026fastdvlm,fu2026nemotronlabsdiffusion} or relax exact verification
\citep{hu2026sdvg,niu2026realtimevla}. The general applicability of L2 drafting
for frozen multimodal targets therefore remains an open question.

We investigate this question through a controlled empirical study of
diffusion-based parallel drafting on multimodal models. Interestingly,
directly transferring an L2 drafter trained on Qwen3 leads to limited
draft--target agreement, while the same block-parallel drafting paradigm
achieves substantially stronger alignment on newer multimodal variants. 
This observation suggests that multimodality itself is not the fundamental
obstacle; instead, the effectiveness of L2 drafting depends on how multimodal
information is represented, aligned, and utilized during generation.
Therefore, rather than asking whether L2 drafting works for multimodal models,
we study under what model, task, and system conditions block-parallel drafting
can translate draft quality improvements into practical acceleration.

In this paper, we present a systematic empirical study of multimodal L2
speculative decoding readiness. Our contributions are:

\begin{enumerate}

    \item \textbf{A unified taxonomy and empirical study of multimodal L2
    drafting.}
    We establish an L0--L2 taxonomy based on drafter-side parallelism and
    evaluate block-parallel drafting across representative multimodal models
    with different scales and architectures.

    \item \textbf{Analysis of multimodal conditioning bottlenecks.}
    We investigate how visual information affects draft--target agreement and
    end-to-end efficiency, revealing the trade-off between multimodal
    awareness and additional conditioning overhead.

    \item \textbf{A readiness characterization for multimodal block-parallel
    decoding.}
    We analyze the impact of model characteristics, task and input
    conditions, and framework support on practical L2 acceleration, and
    summarize remaining challenges toward broader multimodal speculative
    decoding deployment.

\end{enumerate}

The remainder of this paper is organized as follows. Section~\ref{sec:mm_sd}
introduces multimodal speculative decoding and the information-preservation
requirements behind draft verification. Section~\ref{sec:parallel_drafting}
presents our L0--L2 taxonomy of drafter-side parallelism. Section
~\ref{sec:experiments} evaluates the readiness of block-parallel drafting for
multimodal targets through cross-model evaluation, conditioning analysis, and
framework support analysis. Section~\ref{sec:readiness} synthesizes the
empirical findings and discusses future directions before concluding in
Section~\ref{sec:conclusion}.

\section{Multimodal Speculative Decoding}
\label{sec:mm_sd}

\subsection{Foundations}
\label{sec:foundations}

\paragraph{Standard speculative decoding.}
Speculative decoding accelerates autoregressive generation with a draft-and-verify loop. It uses a lightweight drafter to propose multiple future tokens, allowing the target model to verify them in a single forward pass instead of decoding them one by one.
Given a prefix $x_{<t}$, the drafter proposes $K$ future tokens, and the target
scores the drafted positions in parallel while also producing the next-token
distribution after the draft block. Under exact speculative sampling, for
$i=0,\ldots,K-1$, a drafted token $\tilde{x}_{t+i}$ is accepted with
probability
\begin{equation}
    a_i(\tilde{x}_{t+i})
    =
    \min\left\{
        1,\,
        \frac{
            p(\tilde{x}_{t+i}\mid x_{<t},\tilde{x}_{t:t+i-1})
        }{
            q(\tilde{x}_{t+i}\mid x_{<t},\tilde{x}_{t:t+i-1})
        }
    \right\},
    \label{eq:acceptance}
\end{equation}
where $\tilde{x}_{t:t-1}$ is empty for $i=0$. At the first rejection, the
accepted prefix is retained and a correction token is sampled from the
normalized positive residual between the corresponding target and draft
distributions; after a full acceptance, the target contributes one additional
token.
This procedure preserves the target sampling
distribution~\citep{leviathan2023fast,chen2023accelerating}. Its benefit
depends on whether the accepted output amortizes the cost of drafting and
verification. Moreover, although the target evaluates multiple drafted
positions in parallel, a conventional drafter produces them in causal order,
so draft generation can remain a serial bottleneck.

\paragraph{Sequential and parallel drafting.}
Parallel drafting reduces this draft-side dependence rather than merely
parallelizing target verification. It can take different forms, from predicting
multiple future positions in one pass to generating or refining a future block
as a coupled state~\citep{stern2018blockwise,cai2024medusa,
gloeckle2024better}. DFlash exemplifies diffusion-based parallel drafting by
using a lightweight block-diffusion model to generate a future block for
verification by an autoregressive target~\citep{chen2026dflash}.
Section~\ref{sec:parallel_drafting} formalizes the progression from sequential
to increasingly parallel draft generation. These mechanisms are commonly
studied for text generation, but multimodal settings introduce additional
constraints on both input processing and output verification.

\paragraph{Multimodal speculative decoding.}
Unlike text-only SD, a multimodal drafter may need to process image, video,
audio, or environment information that also conditions the target. Access to
this information can improve target--draft agreement, but introduces additional
modality-specific computation and memory cost. Some methods instead reuse
target-derived features or draft without directly processing the multimodal
input. Multimodal applications may also generate actions, acoustic or codec
tokens, and visual tokens or latents rather than text tokens. For these outputs,
exact token identity is not always the only useful notion of validity, and some
methods adopt task- or representation-specific acceptance rules. The next
subsection introduces the analytical dimensions used to compare these design
choices across application domains.




\subsection{Multimodal Speculative Decoding Landscape}
\label{sec:mm_landscape}

We organize the multimodal speculative decoding literature considered in this
survey into five application domains: vision--language models,
video--language models, vision--language--action models, speech and audio
models, and autoregressive visual generation. These domains differ in the
multimodal information available during drafting, the outputs being generated,
and their task and deployment constraints. Such differences affect both how
speculative candidates are produced and how they can be verified, motivating
application domain as the primary organization of the review.

Within each domain, we examine how candidate generation varies in its serial
dependence across future positions. We annotate this property using the
L0--L2 scale, which is formalized in
Section~\ref{sec:parallel_drafting}. Methods with the same P-level can still
differ substantially in how the drafter accesses multimodal information, how
candidates are organized or obtained, how they are accepted or verified, and
whether speculative computation adapts to the current input or state. We track
these properties separately because they can change the effectiveness of
speculative inference without changing future-position dependence.
Table~\ref{tab:mm_sd_landscape} summarizes these method-level characteristics,
while the discussion below synthesizes the main design patterns and available
evidence within each application domain.

\begin{table*}[t]
    \centering
    \caption{Taxonomy of representative multimodal speculative decoding methods
    across application domains, drafter designs, and speculative mechanisms.}
    \label{tab:mm_sd_landscape}
    \scriptsize
    \setlength{\tabcolsep}{2.2pt}
    \renewcommand{\arraystretch}{1.1}
    \begin{tabularx}{\textwidth}{
        @{}
        >{\raggedright\arraybackslash}p{1.42in}
        >{\centering\arraybackslash}p{0.38in}
        >{\centering\arraybackslash}p{0.91in}
        >{\raggedright\arraybackslash}p{0.82in}
        >{\raggedright\arraybackslash}p{0.88in}
        >{\raggedright\arraybackslash}X
        @{}
    }
        \toprule
        \textbf{Method}
        & \multicolumn{2}{c}{\textbf{Drafter}}
        & \multicolumn{3}{c}{\textbf{Mechanism}} \\
        \cmidrule(lr){2-3}\cmidrule(l){4-6}
        & \textbf{L}
        & \textbf{Condition}
        & \textbf{Candidate form}
        & \textbf{Acceptance}
        & \textbf{Adaptation} \\
        \midrule

        \multicolumn{6}{@{}l}{\textit{\textbf{Vision--Language Models}}} \\
        SPD-MLLM~\citeyearpar{gagrani2024multimodal}
        & L0 & Text / direct & Chain & Exact & Static \\
        MASSV~\citeyearpar{ganesan2025massv}
        & L0 & Target feat. & Chain & Exact & Static \\
        DREAM~\citeyearpar{hu2025dream}
        & L0 & Target feat. & Chain & Exact & Static \\
        HiViS~\citeyearpar{xie2026hivis}
        & L0 & Target feat. & Chain & Exact & Static \\
        SpecVLM (image)~\citeyearpar{huang2025specvlm}
        & L0 & Direct / target feat. & Chain / tree & Exact & Condition \\
        ViSpec~\citeyearpar{kang2025vispec}
        & L0 & Direct & Tree & Exact & Candidate \\
        TIGER~\citeyearpar{vo2026tiger}
        & L0 & Direct & Chain & Exact & Condition \\
        Spec-LLaVA / SAGE~\citeyearpar{huo2025specllava,tong2026sage}
        & L0 & Direct & Tree & Exact & Candidate \\
        ViSkip~\citeyearpar{shen2026mmspec}
        & -- & Target feat. & Inherited & Inherited & Execution \\
        SpecFLASH~\citeyearpar{wang2025flash}
        & L1 & Direct / target feat. & Parallel offsets & Exact & Static \\
        Fast-dVLM~\citeyearpar{wu2026fastdvlm}
        & L2 & Direct & Joint block & Exact (greedy) & Static \\
        Nemotron-Labs-Diffusion-VLM~\citeyearpar{fu2026nemotronlabsdiffusion}
        & L2 & Direct & Joint block & Exact (greedy) & Static \\

        \addlinespace[1.5pt]
        \multicolumn{6}{@{}l}{\textit{\textbf{Video--Language Models}}} \\
        Sparse-to-Dense~\citeyearpar{zhang2025sparse}
        & L0 & Direct & Chain & Exact & Candidate \\
        SpecVLM (video)~\citeyearpar{ji2025videospecvlm}
        & L0 & Direct / target feat. & Tree & Exact & Condition \\
        Sparrow~\citeyearpar{zhang2026sparrow}
        & L0 & Target feat. & Tree & Exact & Static \\
        HIPPO / ParallelVLM~\citeyearpar{lv2026hippo,kong2026parallelvlm}
        & L0 & Direct / target feat. & Chain & Exact & Condition / execution \\
        LVSpec~\citeyearpar{ji2026lvspec}
        & L0 & Direct & Chain & Relaxed (position) & Verification \\

        \addlinespace[1.5pt]
        \multicolumn{6}{@{}l}{\textit{Axis 1: Vision--Language--Action Models}} \\
        Spec-VLA~\citeyearpar{wang2025specvla}
        & L0 & Direct & Chain & Relaxed (distance) & Static \\
        HeiSD~\citeyearpar{zheng2026heisd}
        & L0 & Direct & Chain / retrieval & Relaxed (sequence) & Candidate \\
        Reasoning-aware SD (diff.)~\citeyearpar{dinh2026reasoningaware}
        & L2 & Target feat. & Reasoning block & Exact (greedy) & Static \\
        Realtime-VLA FLASH~\citeyearpar{niu2026realtimevla}
        & L2 & Direct & Action block & Consistency & Execution \\

        \addlinespace[1.5pt]
        \multicolumn{6}{@{}l}{\textit{\textbf{Speech and Audio Models}}} \\
        Speech SD~\citeyearpar{lin2025speechsd}
        & L0 & Text / direct & Chain & Relaxed (tolerance) & Static \\
        SpecASR~\citeyearpar{wei2025specasr}
        & L0 & Direct & Tree & Exact & Candidate \\
        ParaASR~\citeyearpar{lin2026paraasr}
        & L1 & Direct & Parallel offsets & Exact & Static \\
        VADUSA~\citeyearpar{li2025vadusa}
        & L1 & Text / direct & Offsets / tree & Relaxed (tolerance) & Static \\
        CTC-SSD~\citeyearpar{saon2026ctcssd}
        & L1$^{*}$ & Direct & Frame hypothesis & Confidence-gated & Verification / execution \\
        PCG~\citeyearpar{yanuka2026pcg}
        & L0 & Text / direct & Chain & Group-exact & Static \\

        \addlinespace[1.5pt]
        \multicolumn{6}{@{}l}{\textit{\textbf{Autoregressive Visual Generation}}} \\
        LANTERN~\citeyearpar{jang2025lantern}
        & L0 & Text / direct & Chain & Relaxed (latent) & Static \\
        LANTERN++~\citeyearpar{park2025lantern++}
        & L0 & Text / direct & Tree & Relaxed (latent) & Static \\
        Spatial SD~\citeyearpar{xiang2026ssd}
        & L1 & Text & Parallel rows & Correction-based & Static \\
        SDVG~\citeyearpar{hu2026sdvg}
        & L2 & Text & Video block & Quality-gated & Execution \\
        Speculative Jacobi~\citeyearpar{teng2025sjd}
        & L2$^{*}$ & Text / direct & Iterative block & Convergence & Static \\
        \bottomrule
    \end{tabularx}

    \vspace{1pt}
    \parbox{\textwidth}{\scriptsize
    \textit{Notation.} ``Direct'' denotes direct non-text conditioning; ``target
    feat.'' denotes reused target features. P-level measures dependence across
    future draft positions, independent of candidate form: L0 = sequential,
    L1 = multi-position parallel prediction, L2 = block-parallel generative.
    L1$^{*}$ parallelizes frame labels; L2$^{*}$ denotes target self-refinement
    (block-parallel but not an independent drafter). ``Static'' indicates no
    runtime adaptation. ViSkip inherits the wrapped method's attributes;
    ``group-exact'' preserves the induced group variable. Rows are representative,
    not exhaustive.}
\end{table*}

\paragraph{Vision--language models.}

In VLMs, visual access can improve target--draft alignment, but its processing
cost can offset the latency saved by accepting more draft tokens. Early work
therefore asks whether the drafter needs direct visual input at all. SPD-MLLM
shows that a language-only drafter can already accelerate a visually
conditioned target, while adding a compact image adaptor produces
task-dependent rather than uniform gains~\citep{gagrani2024multimodal}. MSD
instead develops an explicitly multimodal drafter: it processes text and visual
tokens separately and uses text-only followed by multimodal training to acquire
both language-modeling and visual-perception capabilities~\citep{lin2025msd}.
Both retain an autoregressive draft trajectory, but establish the two endpoints
between condition-blind and explicitly multimodal drafting.

Subsequent methods seek less redundant or more selective access to visual
information. MASSV reuses target vision-encoder outputs through a lightweight
projector~\citep{ganesan2025massv}; Huang et al.'s SpecVLM employs an elastic
visual compressor~\citep{huang2025specvlm}; and ViSpec combines a compact
vision adaptor with persistent global-feature injection~\citep{kang2025vispec}.
Other methods increasingly reuse representations already computed by the
target. At inference time, DREAM cross-attends to cached target features while
compressing the visual context; during training, it uses entropy-guided
selection of intermediate target representations~\citep{hu2025dream}. HiViS
removes visual tokens from the drafter input and transfers visual semantics
through target-derived hidden representations~\citep{xie2026hivis}. TIGER
makes this access step-dependent by routing visual tokens according to the
current textual state and trains the drafter with a verifier-derived
accepted-prefix reward~\citep{vo2026tiger}. This progression moves from
duplicating explicit visual inputs toward compressed, reused, and selectively
exposed visual information.

Candidate topology and runtime control provide complementary improvements
without necessarily changing future-position dependence. ViSpec also uses a
dynamic draft tree, while Spec-LLaVA expands and prunes tree branches according
to draft confidence~\citep{huo2025specllava} and SAGE adapts tree depth and
width using online entropy~\citep{tong2026sage}. ViSkip instead uses
vision-aware signals to adapt when speculative decoding is
attempted~\citep{shen2026mmspec}. Trees broaden the candidate space, whereas
ViSkip controls the allocation of speculative computation; neither mechanism
alone removes the serial dependence of the underlying drafter.

SpecFLASH changes candidate generation more directly. Its semi-autoregressive
drafter processes $K$ placeholder tokens and predicts the next $K$ token
distributions in one pass~\citep{wang2025flash}. We classify it as L1 because
multiple fixed future offsets are predicted in parallel from a shared accepted
context, rather than generated through a jointly evolving block state.

Block-parallel VLMs provide direct L2 examples. Fast-dVLM converts an
autoregressive VLM into a KV-cache-compatible block-diffusion model and supports
self-speculative block decoding~\citep{wu2026fastdvlm}.
Nemotron-Labs-Diffusion jointly trains autoregressive and diffusion objectives
within a shared architecture; its VLM variants can draft through diffusion and
verify autoregressively~\citep{fu2026nemotronlabsdiffusion}. In both cases, the
future block becomes the draft-side generative object rather than a collection
of independently attached offset predictions.

The VLM literature thus spans L0 systems centered on visual access and
candidate coverage, L1 multi-position prediction, and L2 block-parallel
generative drafting. Current L2 evidence primarily comes from co-designed models in which
the same backbone is explicitly trained to support block-parallel drafting and
autoregressive verification. It therefore demonstrates block-level
self-speculation in VLMs, while remaining distinct from an independent,
low-cost parallel drafter that can be paired with heterogeneous pretrained VLM
targets.

\begin{figure}
    \centering
    \includegraphics[width=\linewidth]{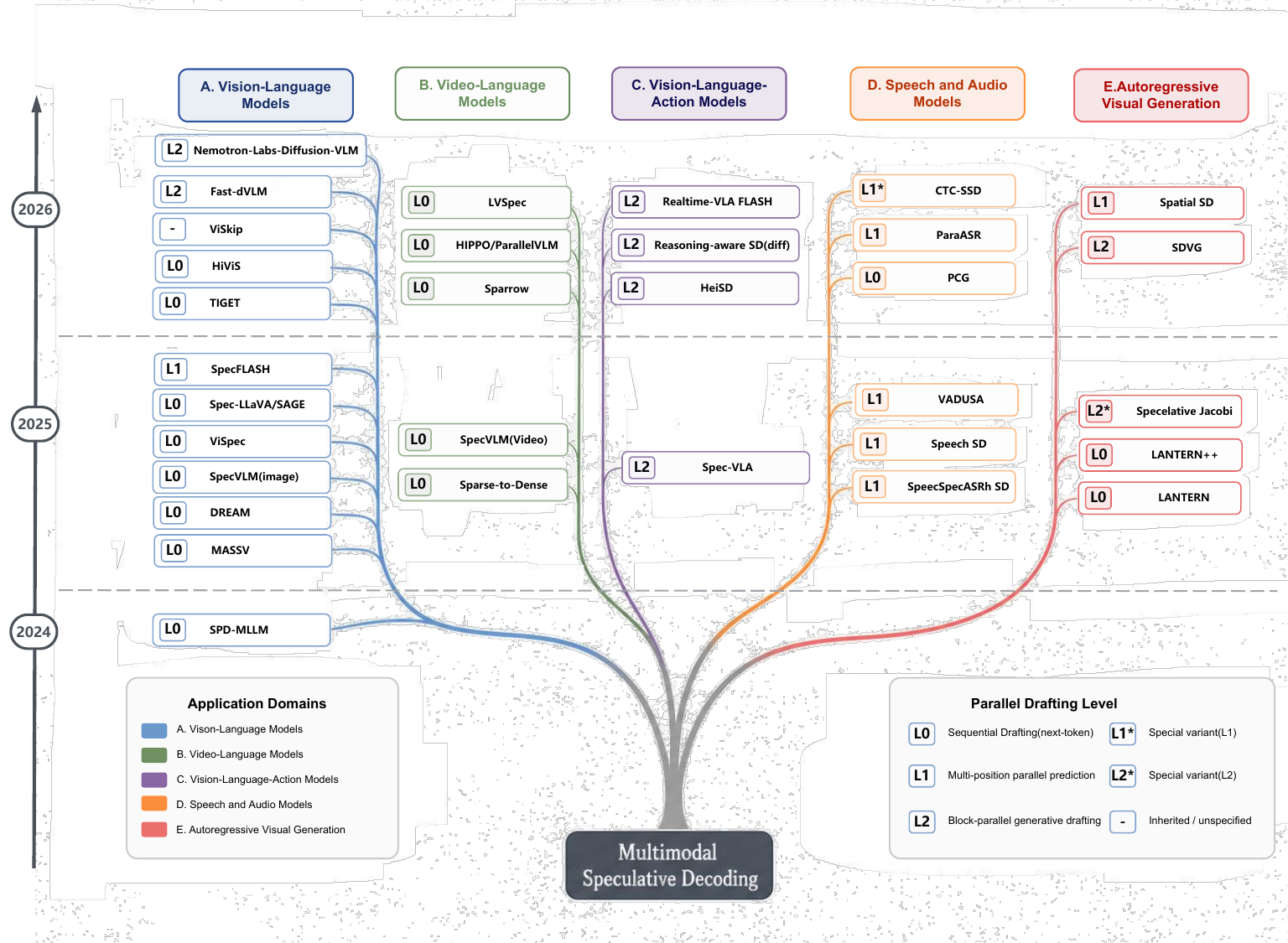}
    \caption{Taxonomy and evolution of representative multimodal speculative decoding methods.}
    \label{fig:struct}
\end{figure}




\paragraph{Video--language models.}

Video-language models amplify the condition-cost problem because long visual
prefixes enlarge draft-side prefill, attention, and KV-cache costs. A small
drafter may also lose accuracy when relevant evidence is diluted across the
video context. Video speculative decoding has consequently focused on reducing
condition-processing cost and overlapping system execution, while candidate
generation remains largely autoregressive.

Sparse-to-Dense (StD) reduces draft computation without changing the input
representation: it uses sparse top-$K$ attention during drafting and restores
dense attention for target verification~\citep{zhang2025sparse}. SpecVLM
instead sparsifies the condition presented to the drafter through
verifier-guided and spatially uniform token pruning, while the target verifies
against the full video context~\citep{ji2025videospecvlm}. The two systems thus
apply sparsity to different objects---draft attention and draft input---while
retaining strict target-token verification.

Sparrow questions whether the drafter should process video tokens at all. It
attributes long-video degradation to visual KV-cache growth, context mismatch,
and attention dilution, and instead reuses target-side text states that have
already integrated visual information~\citep{zhang2026sparrow}. Its recursive
multi-token prediction exposes the drafter to predicted states during training;
the deployed draft trajectory remains autoregressive, optionally organized as
a token tree.

A separate line reduces mutual waiting between draft and target models. HIPPO
combines spatial--temporal relevance pruning with an execution policy that
overlaps the next proposal and current verification, adapting between
optimistic and conservative schedules from recent
outcomes~\citep{lv2026hippo}. ParallelVLM similarly combines verifier-guided
pruning with overlapped execution and corrects positional bias in
attention-based token selection~\citep{kong2026parallelvlm}. This pipeline
overlap is systems parallelism rather than parallel generation of future draft
positions.

LVSpec modifies the acceptance contract instead. It applies strict verification
to visually grounded anchors, relaxes acceptance for less visually relevant
tokens, and allows nearby-position matching between target and draft
sequences~\citep{ji2026lvspec}. Because this rule need not reproduce the exact
target sequence, its reported preservation of more than $99.8\%$ of target task
performance is empirical high-fidelity evidence rather than a formal exactness
guarantee.

Current video-specific methods are therefore concentrated at L0 but obtain
acceleration from three distinct sources: condition sparsification or feature
reuse, draft--verify pipeline overlap, and visually informed relaxed
acceptance. Block-diffusion VLMs do not yet resolve the long-video setting,
where a parallel drafter must avoid recreating the target's condition cost
while preserving temporal grounding and competing with L0 systems that already
prune, reuse, or hide draft computation.




\paragraph{Vision--language--action models.}

VLA systems place speculative inference inside a closed-loop control pipeline,
which creates two verification problems. Under a fixed observation, exact
equality between discretized action tokens may be unnecessarily strict because
nearby tokens can encode functionally similar controls. Across time, however,
even an initially valid action chunk can become unsafe or obsolete as new
observations arrive. Verification must therefore address both candidate--target
agreement and the continuing validity of an accepted plan.

Spec-VLA studies the first problem with a compact autoregressive VLA drafter and
accepts a proposed action token when its discretized-bin distance from the
target token is below a threshold~\citep{wang2025specvla}. KERV adds
kinematics-aware rectification: a Kalman predictor repairs speculative errors,
while a runtime signal adjusts the acceptance threshold~\citep{zheng2026kerv}.
HeiSD instead routes geometrically irregular segments to a learned VLA drafter
and regular segments to retrieved action sequences, using verify-skip and
sequence-wise relaxed acceptance on the retrieval path~\citep{zheng2026heisd}.
The learned draft paths remain autoregressive, while retrieval supplies a
different candidate source. Correctness is consequently supported by task
success and trajectory quality rather than exact reproduction of the target
action sequence.

SV-VLA addresses the second problem explicitly. A heavy VLA acts as a
low-frequency macro-planner that emits an action chunk and planning context,
whereas a lightweight verifier repeatedly compares the planned action with a
closed-loop reference conditioned on the latest observation and triggers
replanning when needed~\citep{wang2026svvla}. This is not merely a looser
token-matching rule: it tests whether an accepted open-loop plan remains valid
after the environment and execution history have changed. Its contribution
therefore lies in verification and runtime adaptation rather than in a new
candidate-generation P-level.

Recent work also changes the object generated in parallel. Reasoning-aware
speculative decoding accelerates the autoregressive chain of causation that
precedes trajectory prediction. Its block-diffusion variant is L2 because it
proposes a future reasoning block jointly, although the parallelized object is
language reasoning rather than control~\citep{dinh2026reasoningaware}.
Realtime-VLA FLASH provides direct action-side evidence: a lightweight model
drafts a continuous action chunk, the target Action Expert verifies the chunk
in parallel, and a phase-aware mechanism falls back to full inference when
consistency is insufficient~\citep{niu2026realtimevla}. We record this joint
action-chunk proposal as L2, with consistency-based rather than exact
verification. Fast-dDrive supplies a complementary L2 design for autonomous
driving. Its block-diffusion VLA refines tokens within scaffolded semantic
sections, including a trajectory section, and uses scaffold speculative
decoding to accelerate the structured output~\citep{zhang2026fastddrive}.

L2 evidence within the VLA domain thus spans distinct generated objects:
reasoning blocks, continuous action chunks, and structured outputs containing
trajectories. The first parallelizes upstream language reasoning, whereas the
latter two place speculative generation directly on action or trajectory
outputs. These control-side settings must additionally validate accepted plans
under evolving observations, provide safe fallback and recovery, and establish
end-to-end gains across control tasks and VLA architectures.




\paragraph{Speech and audio models.}

Speech speculation spans three output settings. ASR produces a transcript that
is strongly constrained by the acoustic input; audio-language systems generate
more open-ended text; and speech synthesis emits long acoustic- or codec-token
sequences for which different tokenizations may decode to perceptually similar
waveforms. These settings offer different sources of predictability and require
different notions of verification.

For ASR, SpecASR uses a smaller audio-conditioned model with adaptive draft
length, sequence recycling, and sparse-tree verification~\citep{wei2025specasr}.
Its learned draft trajectory remains L0; adaptation and tree width reduce wasted
work without changing its serial depth. ParaASR provides direct L1 evidence by
training multiple future-transcript-token branches on an audio-language model.
It proposes six transcript tokens per decoding step and commits only the prefix
verified by the autoregressive head~\citep{lin2026paraasr}. CTC-based proposals
parallelize a different unit. CTC-SSD reuses the acoustic encoder to predict
frame labels simultaneously, collapses them into a complete transcript
hypothesis, and either accepts a low-entropy hypothesis, verifies its token
likelihoods in one LLM pass, or resumes autoregressive decoding from the
accepted prefix~\citep{saon2026ctcssd}. We denote this as L1$^{\ast}$: it removes
output-length-dependent drafting, as L1 intends, but parallelizes acoustic
frames rather than directly predicting future transcript positions. SMUD uses
a related CTC preliminary transcript and batched decoder comparison to locate
regions that still require causal search~\citep{okabe2025smud}.

Speech synthesis shifts attention from transcript alignment to perceptual
equivalence. Speech Speculative Decoding (Speech SD) employs a lightweight L0
speech LM and adds a heuristic tolerance factor to its sampling acceptance
probability~\citep{lin2025speechsd}. VADUSA instead attaches Medusa-style heads
for parallel future speech-token prediction, forms sparse candidate trees, and
uses tolerance-based verification; the heads make it a direct L1 speculative
TTS design~\citep{li2025vadusa}. Codec-MTP is also L1, but uses several
future-token heads with Viterbi path selection rather than longest-prefix tree
verification~\citep{nguyen2025codecmtp}. These relaxed or path-level procedures
are supported by intelligibility, naturalness, and speaker-similarity evidence,
not by preservation of the target token distribution. Principled
Coarse-Graining (PCG) gives this distinction a more formal treatment: it forms
overlapping Acoustic Similarity Groups from the target embedding space and
performs rejection sampling over the induced group
distribution~\citep{yanuka2026pcg}. Its exactness guarantee applies to the group
variable, not to the target's original token sequence, so perceptual evaluation
remains necessary when substituting tokens within a group.

For open-ended audio-conditioned text, UGSD drafts speech-emotion captions on
an edge model and escalates only uncertain blocks to a cloud verifier, making
runtime adaptation trade off latency, communication, and
privacy~\citep{xue2026ugsd}.

Adjacent block-generation work provides enabling evidence for speech-side L2.
Chatterbox-Flash is not itself a draft--verify speculative system: it converts
an autoregressive TTS decoder into a block-diffusion decoder that generates
speech tokens in parallel within streaming
blocks~\citep{seo2026chatterboxflash}. It therefore demonstrates an enabling
block-generation capability rather than a complete L2 speculative pipeline.
What remains unresolved is whether such a generator can serve as a sufficiently
cheap proposal model for an autoregressive target, and how its blocks should be
verified under an exact or perceptually justified contract without erasing the
latency saved by parallel drafting.




\paragraph{Autoregressive visual generation.}

In autoregressive visual generation, the speculative object is a spatial or
spatiotemporal representation, including discrete image codes, continuous
latents, and video blocks. Exact identity may be unnecessarily restrictive
when several codes decode to similar content, yet a local substitution can
propagate through later regions. The literature therefore explores both
alternative candidate-construction mechanisms and output-aware verification;
only some methods parallelize proposal generation.

For discrete visual tokens, LANTERN accepts candidates over neighborhoods in
the latent codebook and bounds the resulting distributional shift in total
variation~\citep{jang2025lantern}. LANTERN++ combines this relaxed contract
with a static tree to avoid shallow drafts under low-confidence visual
distributions~\citep{park2025lantern++}. Both retain a L0 autoregressive
proposal trajectory; their main innovations lie in visual equivalence and
candidate coverage rather than future-position parallelism.

Geometry-aware parallel decoding provides adjacent evidence. ZipAR is a
target-side parallel decoder that predicts row- and column-direction tokens
concurrently using adaptive local windows and rejection sampling analogous to
speculative decoding~\citep{he2025zipar}; it is not a conventional
drafter--target speculative system. Spatially Speculative Decoding (Spatial
SD), by contrast, learns lightweight horizontal and vertical latent heads that
draft subsequent rows in parallel, followed by target-side
auto-correction~\citep{xiang2026ssd}. These fixed-offset heads constitute L1:
they predict many positions per pass without jointly modeling the entire
future block. MuLo-SD supplies a complementary coarse-to-fine design, using a
low-resolution drafter and restricting rejection and resampling to local
neighborhoods rather than discarding the full raster-order
suffix~\citep{peruzzo2026mulosd}. Beyond discrete codes, Continuous
Speculative Decoding derives acceptance and rejection procedures for the
diffusion distributions of continuous-valued visual AR models, with trajectory
alignment improving draft--target agreement~\citep{wang2024continuous}. VVS
instead reduces target cost by dynamically skipping selected verification
rounds and reusing cached token features~\citep{dong2026vvs}.

Speculative Jacobi Decoding (SJD) follows a drafter-free route: the target
iteratively updates a future token window and commits a prefix under a
probabilistic convergence criterion~\citep{teng2025sjd}. We annotate it as
L2$^{\ast}$ because it refines a parallel block but does not use a separate
low-cost drafter (the asterisk denotes block-parallel refinement without an
independent drafter). GSD inherits this SJD trajectory and replaces token-level
verification with relaxed acceptance over dynamically constructed groups of
visually valid codes~\citep{so2025gsd}; its contribution therefore lies
primarily in the acceptance contract. SJD2 and SJD++ respectively stabilize
refinement through next-clean-token prediction and high-confidence token
retention~\citep{teng2025sjd2,teng2025sjdpp}, while Speculative Coupled
Decoding couples successive samples to improve stability without changing the
target sampling distribution~\citep{so2026scd}.


\begin{table*}[t]
    \centering
    \caption{Cross-domain synthesis of multimodal speculative decoding.
    L-level entries report representative candidate-generation mechanisms
    rather than exhaustive coverage within each domain.}
    \label{tab:mm_cross_domain}
    \small
    \setlength{\tabcolsep}{6pt}
    \renewcommand{\arraystretch}{1.15}
    \begin{tabularx}{\textwidth}{
        @{}
        >{\raggedright\arraybackslash}p{1in}
        >{\raggedright\arraybackslash}p{1in}
        >{\centering\arraybackslash}p{1in}
        >{\raggedright\arraybackslash}p{1in}
        >{\raggedright\arraybackslash}X
        @{}
    }
        \toprule
        \textbf{Application}
        & \textbf{Domain constraint}
        & \textbf{Draft profile}
        & \textbf{Acceptance semantics}
        & \textbf{Domain-specific cue} \\
        \midrule

        \rule[-2.1ex]{0pt}{5.0ex}VLM
        & Grounding--cost trade-off
        & L0 / L1 / L2
        & Token exactness
        & Visual relevance \\

        \rule[-2.1ex]{0pt}{5.0ex}Video-LM
        & Long-context cost
        & L0
        & Token exactness
        & Temporal relevance \\

        \rule[-2.1ex]{0pt}{5.0ex}VLA
        & Closed-loop state
        & L0 / L2
        & Functional validity
        & Control phase \\

        \rule[-2.1ex]{0pt}{5.0ex}Speech/Audio
        & Temporal dependencies
        & L0 / L1
        & Sequence / perceptual
        & Prediction uncertainty \\

        \rule[-2.1ex]{0pt}{5.0ex}Visual Generation
        & Spatiotemporal coupling
        & L0 / L1 / L2$^{(*)}$
        & Group / quality
        & Spatial locality \\

        \bottomrule
    \end{tabularx}

    \vspace{1pt}
    \parbox{\textwidth}{\small
    \textit{Notation.} L2$^{(*)}$ indicates that both block-level proposals
    and L2$^{*}$ target-side self-refinement are represented.}
\end{table*}

This application domain is also extending to video. SDVG uses a smaller
denoising model to propose continuous spatiotemporal blocks for a
block-autoregressive video target and verifies them with an image-quality
router~\citep{hu2026sdvg}. It is a L2 block-proposal design, but uses a
quality-gated rather than exact distributional contract. Overall, visual
generation spans L0 autoregressive drafting, geometry-aware L1 prediction, L2
block proposals, and L2$^{\ast}$ target self-refinement. Across these forms,
the output-level correctness claim remains inseparable from the reported
efficiency gain.

\subsection{Cross-Domain Analysis}
\label{sec:cross_domain}

Table~\ref{tab:mm_cross_domain} condenses the preceding review along three
cross-domain dimensions: candidate-generation parallelism, acceptance
semantics, and domain-specific cues for adapting speculative computation.

Across domains, multimodal conditioning and output structure change both the
cost profile of speculation and the information available for adapting it.
Domain-specific cues can act on condition access, candidate construction,
verification, or system execution. For example, relevance can determine which
condition features are exposed, while uncertainty can govern whether
speculation or verification should continue. Such adaptations can reduce
condition exposure, redundant computation, or wasted verification without
changing the dependence among future candidate positions. They are therefore
complementary to candidate-generation parallelism rather than evidence of a
higher L-level.

Acceptance contracts vary more directly with the generated object. Text-output
systems can retain token-level target equivalence, whereas action, speech, and
visual outputs also admit functional, sequence-level, group-level, or
perceptual notions of validity. Exact contracts provide target-distribution or
greedy-output equivalence under their stated assumptions. Relaxed contracts
instead require empirical fidelity evidence in addition to any formal bound on
their deviation from the target. This distinction becomes more consequential
for block candidates, whose usefulness depends on whether they can be checked
under an acceptance contract appropriate to the output space.

Across the reviewed domains, candidate generation, condition access,
verification, and runtime adaptation have developed along different paths.
Their effects are coupled in reported systems, leaving unresolved what is
gained---and what new costs arise---when candidate generation moves from serial
proposals to jointly generated or refined blocks. Addressing this question
requires separating future-position dependence from domain-specific
optimizations. Section~\ref{sec:parallel_drafting} therefore formalizes the
progression from sequential drafting to L2 block generation and examines
block-parallel generative drafting within that progression.
Section~\ref{sec:experiments} then evaluates this mechanism empirically in
multimodal inference.


\section{Evolution of Drafter-Side Parallelism}
\label{sec:parallel_drafting}

\subsection{From Sequential to Parallel Drafting}

Early speculative decoding uses a small autoregressive model to propose future
tokens~\citep{leviathan2023fast,chen2023accelerating}. The EAGLE series instead
predicts target-model features and maps them back to token distributions
~\citep{li2024eagle1,li2024eagle2,li2025eagle3}. These methods improve draft
quality and can expand multiple candidate branches, but the underlying draft
trajectory remains sequential: moving to a deeper future position requires
another causally dependent drafter forward.

A different line of work reduces this sequential drafting depth.
Medusa and multi-token prediction predict several future positions in parallel
~\citep{cai2024medusa,gloeckle2024better}. More generally, parallel drafters
allow one drafter forward to advance several future positions rather than only
the next one. Block-parallel generative drafting takes a further structural step.
DFlash and DSpark organize the main drafting computation around an entire
future block, allowing its positions to be processed in parallel
~\citep{chen2026dflash,cheng2026dspark}.

This progression motivates a L0--L2 taxonomy of speculative drafting.
The taxonomy characterizes \emph{drafter-side parallelism}: how far candidate
generation can advance along the future sequence within one drafter forward.
It is independent of how many alternative candidates are retained or how they
are verified.

\subsection{A L0--L2 Taxonomy of Speculative Drafting}

\begin{figure}
    \centering
    \includegraphics[width=\linewidth]{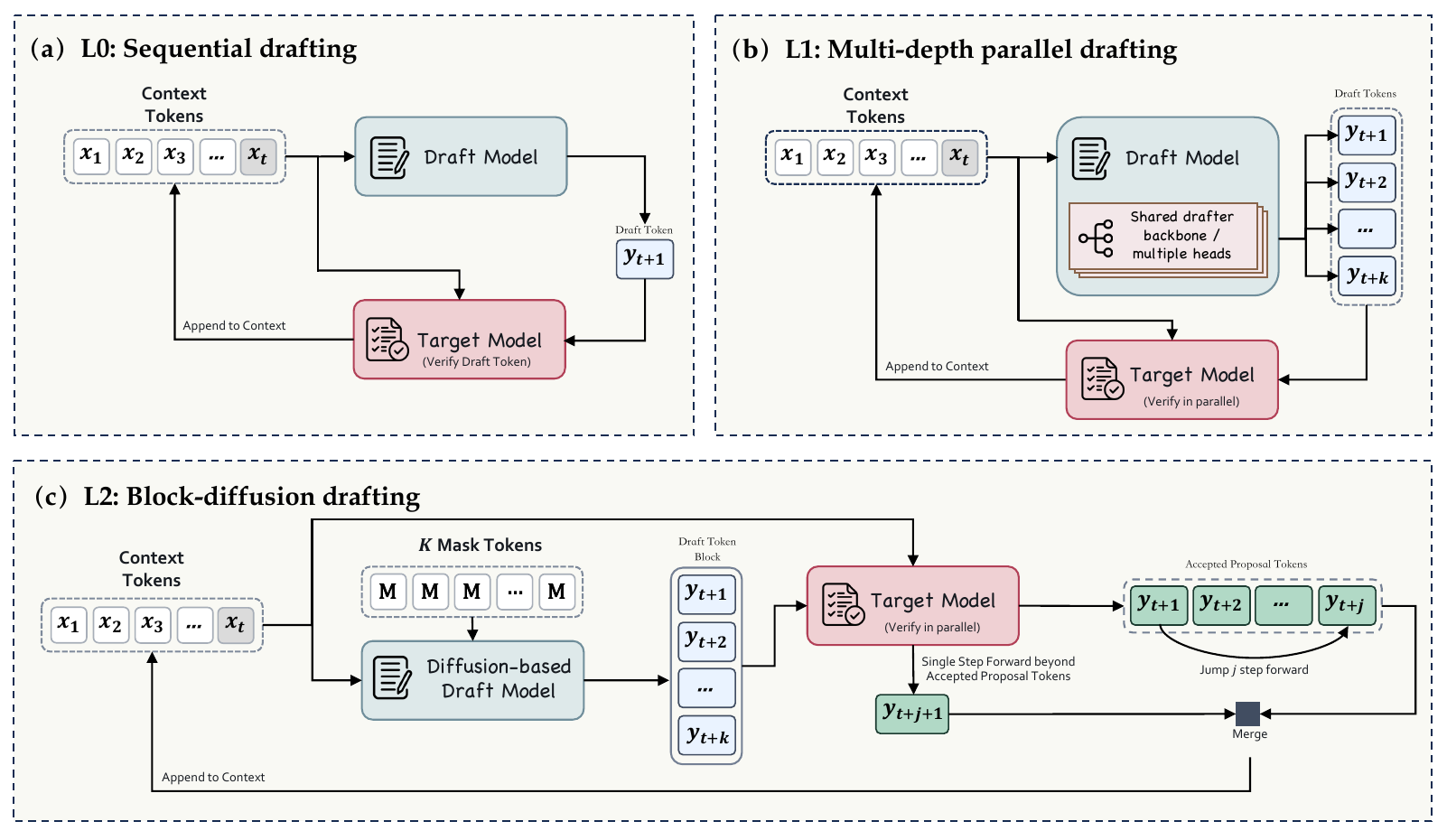}
    \caption{Illustration of the L0–L2 taxonomy of drafter-side parallelism. L0 performs sequential drafting with one future position advanced per drafter forward; L1 predicts multiple predefined future positions in parallel from a shared prefix; and L2 treats the future block as a jointly generated state, enabling block-parallel drafting followed by parallel target verification.}
    \label{fig:struct}
\end{figure}

We first define the terminology used in the taxonomy.
Consider one speculation round that proposes $K$ future positions.
A \emph{candidate trajectory} is an ordered sequence of proposed future tokens.
Its \emph{draft depth} denotes the position along this trajectory: depth $1$
corresponds to the next token, depth $2$ to the following token, and so on.
A \emph{drafter forward} denotes one invocation of the main drafter network.
Finally, \emph{candidate width} denotes the number of alternative candidates
maintained at the same draft depth.

The L0--L2 taxonomy captures a fundamental progression:
token generation $\rightarrow$ parallel position prediction $\rightarrow$ block generation.
L1 parallelizes prediction across future positions; L2 changes the generative unit
from individual future positions to the future block itself.

\paragraph{L0: Sequential drafting.}
L0 advances a candidate trajectory one depth at a time.
To propose $K$ consecutive future positions, the drafter therefore performs
$K$ causally dependent forwards. Small autoregressive language models and the
EAGLE family belong to this regime. EAGLE may produce several alternative
candidates at one depth, but these candidates increase \emph{width}, not
\emph{depth}: reaching the next depth still requires another drafter forward.
Mathematically, L0 produces a sequence through repeated application:
$p(x_{t+1}), p(x_{t+2} \mid x_{t+1}), \ldots, p(x_{t+K} \mid x_{t:t+K-1})$.

\paragraph{L1: Multi-position parallel drafting.}
L1 breaks the one-forward--one-depth constraint by predicting several
predefined future positions in parallel from shared prefix features.
The definition is intentionally architecture-agnostic: parallelism may be
implemented with multiple prediction heads, multi-token prediction, grouped
prediction, semi-autoregressive chunks, or other parallel drafting mechanisms.
What defines L1 is not the specific architecture, but that several future
depths are predicted independently from a shared context rather than generated
as a jointly evolving block.
Mathematically, L1 produces marginals
$p(x_{t+1}), p(x_{t+2}), \ldots, p(x_{t+K})$
from the accepted prefix, without joint block-level state updates.

\paragraph{L2: Block-parallel generative drafting.}
L2 identifies a block-native form of parallel drafting.
Rather than treating future positions only as several predictions to be made
in parallel, the drafter takes the future block itself as the basic generative
unit. The entire block evolves jointly through refinement steps:
$X_{t:t+K}^{(0)} \rightarrow X_{t:t+K}^{(1)} \rightarrow \cdots \rightarrow X_{t:t+K}^{(S)}$,
where $S$ is bounded and does not scale linearly with $K$.
Block diffusion, exemplified by DFlash and DSpark~\citep{chen2026dflash,cheng2026dspark},
is the representative realization considered in this paper, but L2 is not
limited to diffusion architectures. Lightweight dependency modeling inside
the block does not change this classification; L2 is defined by the
block-parallel organization of the main drafting computation.

The distinction between L1 and L2 is therefore structural rather than a simple
count of outputs per forward. Both may advance several future depths in one
forward. L1 predicts several predefined future offsets from a shared prefix,
whereas L2 treats the future block as a jointly evolving generative state.

\begin{table*}[t]
    \centering
    \small
    \setlength{\tabcolsep}{6pt}
    \caption{
    A L0--L2 taxonomy of drafter-side parallelism.
    $K$ denotes the number of future positions considered in one speculation
    round. The taxonomy describes how candidate trajectories are generated;
    candidate width, tree construction, and target-side verification are
    orthogonal.
    }
    \label{tab:parallelism_levels}
    \begin{tabularx}{\textwidth}{c p{1.55in} X X}
        \toprule
        Level &
        Drafting regime &
        Core definition &
        Representative methods \\
        \midrule

        L0 &
        Sequential drafting &
        One forward advances one draft depth; reaching depth $K$ requires
        $K$ causally dependent forwards &
         Small AR LM; \par EAGLE
        \\

        L1 &
        Multi-position parallel drafting &
        One forward predicts several predefined future positions in parallel from
        a shared prefix, but does not treat the future block as a jointly
        evolving generative state &
        Medusa; MTP; \par SpecFLASH
        \\

        L2 &
        Block-parallel generative drafting &
        The future block is the basic generative state; its positions are
        jointly generated or refined with bounded serial drafting depth &
        DFlash;\par DSpark
        \\

        \bottomrule
    \end{tabularx}
\end{table*}

\paragraph{Candidate width is orthogonal to draft depth.}
Draft parallelism should not be confused with candidate branching.
Candidate width measures how many alternatives are retained at a given depth;
draft depth measures how far the candidate trajectory extends into the future.
Tree-based speculative decoding increases the former.
SpecInfer introduces token-tree verification~\citep{miao2024specinfer},
Sequoia optimizes tree topology~\citep{chen2024sequoia}, and DDTree constructs
a budgeted tree from block-parallel predictions~\citep{ringel2026ddtree}.
These mechanisms do not by themselves change the L level of the underlying
drafter. An EAGLE drafter remains L0 when its candidates are organized into a
tree, while DFlash combined with DDTree remains L2.

In short, L0 advances one draft depth per forward, L1 predicts multiple future
positions in parallel, and L2 organizes parallel drafting around an entire
future block as a jointly generative state.

\subsection{The Multimodal L2 Gap} 

Section~\ref{sec:mm_sd} shows that multimodal speculative decoding has advanced rapidly in condition compression, target--drafter alignment, candidate construction, and modality-aware verification. Yet most existing methods remain within L0 or L1 drafting. Evidence for L2---block-parallel generative drafting that treats the future block as the jointly evolving generative state---remains limited. In contrast, text speculative decoding already provides concrete L2 examples such as DFlash and DSpark. Text speculative decoding has entered L2; multimodal speculative decoding largely has not. We call this the \emph{multimodal L2 gap}.

This gap does not imply that multimodal models are inherently unsuitable for block-parallel generative drafting. Multimodal conditioning instead changes the tradeoff. Accessing visual, audio, or other modality features can increase draft cost, while the additional condition may also make future tokens more predictable. A useful multimodal L2 drafter must therefore exploit the condition well enough to produce longer or higher-quality candidate blocks while preserving the efficiency of block-parallel drafting. 

This motivates the controlled experiments in Section~\ref{sec:experiments}. Following prior speculative-decoding surveys ~\citep{xia2024unlocking,hu2025generationrefinement}, we evaluate both draft quality and system efficiency. We report average accepted length, token acceptance rate, target-token rank, and full-block acceptance, together with stage latency, throughput, decode-only and end-to-end speedup, memory, and energy where available.

\section{Is Multimodal Speculative Decoding Ready for L2?}
\label{sec:experiments}

\subsection{Evaluation Questions and Experimental Protocol}
\label{sec:eval_protocol}

Sections~\ref{sec:mm_sd} and~\ref{sec:parallel_drafting} show that multimodal speculative decoding has made
substantial progress in condition compression, candidate construction, and
verification strategies, while the transition from sequential or
multi-position drafting to L2 block-parallel generative drafting remains
unclear. Unlike text-only models, multimodal systems introduce additional
factors, including heterogeneous architectures, expensive modality
conditioning, task-dependent predictability, and system-level overhead.
Therefore, we study L2 readiness through a diagnostic perspective: rather
than asking whether a block-parallel drafter can be constructed, we ask under
which model, task, condition, and system configurations it can translate
draft quality improvements into practical end-to-end acceleration.

We organize our empirical study around four evaluation questions:

\textbf{Q1: Model compatibility.}
Can existing multimodal architectures support effective L2 block-parallel
drafting? 
We evaluate whether lightweight L2 drafters can achieve sufficient
draft--target agreement across different multimodal model families, scales,
and training paradigms. This analysis examines whether L2 readiness is a
general property of multimodal models or depends on specific target
architectures and drafter training strategies.

\textbf{Q2: Conditioning efficiency.}
Does multimodal conditioning remain the dominant bottleneck after draft
generation becomes parallel?
Although L2 drafting substantially reduces the cost of generating future
tokens, multimodal drafters may require additional computation to access
visual or other modality information. We analyze the latency contribution of
conditioning, and investigate whether reducing or removing explicit modality
access can preserve draft quality while improving end-to-end efficiency.

\textbf{Q3: Task and input dependence.}
When do improved draft quality and accepted length translate into practical
acceleration?
Multimodal generation workloads differ significantly in output uncertainty,
grounding requirements, and input complexity. We therefore evaluate L2
drafting across diverse tasks and input conditions to identify when the
accepted-token gain is sufficient to amortize additional speculative
overheads.

\textbf{Q4: System readiness.}
Is the current inference ecosystem mature enough to support practical
multimodal L2 deployment?
Beyond algorithmic effectiveness, practical adoption requires integrated
support for drafter training, candidate generation, verification, and
serving frameworks. We summarize the current ecosystem support for
representative L2 methods and discuss remaining deployment barriers.

Based on these questions, we conduct a cross-model evaluation covering
different multimodal architectures and task categories. We report both draft
quality and system efficiency metrics, including mean accepted tokens (MAT),
token acceptance rate, target-token rank, block acceptance, stage-level
latency, throughput, memory consumption, and end-to-end speedup. Unless
otherwise specified, all experiments use matched decoding configurations and
backend-specific autoregressive baselines to ensure fair comparison.
We use a common protocol to evaluate the L2 transition across model families
and tasks. Numerical cells are deliberately marked ``TBD'' until the final
reproducible export; we do not infer or estimate missing measurements.

\subsection{Experimental Setup}

\paragraph{Models and implementation.}

Our evaluation suite comprises four vision-language models selected to span
diverse configurations along three orthogonal axes: model scale, architectural
family, and pretraining regime. Concretely, we include Qwen3-VL-4B and
Qwen3-VL-8B~\citep{bai2025qwen3vl}, which are natively pretrained as multimodal dense models;
Qwen3.6-27B~\citep{qwen3.6-27b}, a dense model initialized from text-only pretraining and
subsequently aligned to multimodal inputs; and Qwen3.6-35A3B~\citep{qwen36_35b_a3b}, a
Mixture-of-Experts (MoE) model that likewise undergoes text pretraining
followed by multimodal alignment. This selection spans
parameter counts from 4B to 27B (active parameters for MoE) and covers both
end-to-end multimodal pretraining and post-hoc multimodal alignment,
mitigating the risk that conclusions are confounded by a single architectural
choice or training recipe. For targets with publicly released speculative
decoding checkpoints---most notably Qwen3.6-27B---we compare up to five
decoding configurations: (1) target-only autoregressive (AR) decoding, which
serves as the speed and quality baseline; (2) EAGLE-3~\citep{li2025eagle3}
with the PRISM-EAGLE3 drafter checkpoint released by
Ex0bit\footnote{\url{https://huggingface.co/Ex0bit/Qwen3.6-27B-PRISM-EAGLE3}}
(L0); (3) the native multi-token prediction (MTP) heads
~\citep{gloeckle2024better} shipped with Qwen3.6 (L1); (4)
DFlash~\citep{chen2026dflash} with the publicly released z-lab
checkpoint\footnote{\url{https://huggingface.co/z-lab/Qwen3.6-27B-DFlash}} (L2);
and (5) DSpark~\citep{cheng2026dspark} with the publicly released satgeze
checkpoint\footnote{\url{https://huggingface.co/satgeze/Qwen3.6-27B-DSpark}}
(L2). For targets without publicly available L2 checkpoints---in particular
Qwen3-VL-8B---we additionally train a DFlash-style block-parallel drafter
using the SpecForge training framework~\citep{li2026specforge}, ensuring that
our evaluation is not limited to targets for which pre-trained drafters
happen to be publicly released.

Unless otherwise noted (see $^{\dagger}$ in Table~\ref{tab:main_results}), all
experiments are conducted on the SGLang backend under matched greedy decoding
settings, with a maximum generation length of 2,048 new tokens and a fixed
random seed of 42. All methods employ identical prompts, tokenizers, stopping
criteria, and batch sizes. We report end-to-end wall-clock speedup, which
encompasses vision encoding, prefilling, drafter candidate generation, tree
construction, and target model verification.


\paragraph{Tasks and evaluation matrix.}
\label{exp:tasks}

We construct a 600-sample evaluation set following the task balance of MMSpec~\citep{shen2026mmspec}, a dedicated benchmark for multimodal speculative decoding. Concretely, we randomly sample 100 instances from each of six task categories using a fixed random seed of 42: general VQA from GQA~\citep{ainslie2023gqa}, image captioning from Flickr30K~\citep{plummer2015flickr30k}, text-focused VQA from TextVQA~\citep{singh2019towards-textvqa}, chart understanding from CharXiv~\citep{wang2024charxiv}, complex multimodal reasoning from MMMU~\citep{yue2024mmmu}, and multi-turn conversation from the combined ConvBench~\citep{liu2024convbench} and MM-MT-Bench~\citep{agrawal2024pixtral-mm-mt-bench} splits.
The only deviation from MMSpec is that we replace the COCO captioning split with Flickr30K to reduce lexical overlap with common VLM pre-training corpora.


\paragraph{Baselines and metrics.}

The primary baseline is target-only autoregressive decoding executed within the
same inference backend and under identical decoding settings. Candidate quality
is measured by mean accepted tokens (MAT), defined as the average number of
consecutive draft tokens accepted by the target per speculation round. The
primary efficiency metric is end-to-end wall-clock speedup, which encompasses
vision encoding, target prefill, drafter prefill, candidate generation, tree
construction, target verification, and token sampling. Speedup is computed
against the matched target-only baseline:
\begin{equation}
    \mathrm{Speedup}=\frac{T_{\mathrm{AR}}}{T_{\mathrm{method}}}.
\end{equation}

\medskip
The experiments that follow answer the four questions through three empirical
analyses: cross-model effectiveness evaluation, conditioning and input-scaling
analysis, and ecosystem support analysis.

\subsection{Does L2 Block-Parallel Drafting Deliver Multimodal Speedups?}

We begin with the foundational effectiveness question: does L2 block-parallel
drafting brings the acceptance-length and latency gains observed in
text-only speculative decoding to multimodal autoregressive models, or does
the additional draft-side computation and multimodal conditioning overhead
negate these benefits? 
To guard against conclusions that are artifacts of a
single architecture, we evaluate across the four-model suite described above,
which jointly varies model scale (4B--27B total parameters), architectural
family (dense vs.\ MoE), and pretraining regime (native multimodal vs.\
text-pretrained then multimodal aligned). 
For Qwen3-VL-8B, where no public L2 drafter checkpoint is available, 
we train a DFlash-style block-parallel
drafter via SpecForge~\citep{li2026specforge} to avoid selection bias toward
targets with pre-existing third-party checkpoints. 
We complement these head-to-head comparisons with a general bench collected in~\ref{exp:tasks} ---to identify the specific conditions under which L2 drafting succeeds or fails.

\begin{table*}[t]
    \centering
    \setlength{\tabcolsep}{3.2pt}
    \renewcommand{\arraystretch}{1.5}
    \caption{Speculative decoding performance with a maximum of 2,048 new
    tokens. MAT denotes the mean number of output tokens advanced per
    speculative step.
    Overall results are
    task-equal means. The best and second-best speculative results for each
    target are shown in bold and underlined, respectively.
    $^{\dagger}$Results measured on the HuggingFace Transformers backend; all other results use the SGLang backend. Each speedup is normalized to its own backend-matched autoregressive baseline. We abbreviate target models Q denotes Qwen.zheli}
    \vspace{6pt}
    \label{tab:main_results}
    \resizebox{\textwidth}{!}{%
    \begin{tabular}{ll*{7}{cc}}
        \toprule
        \multirow{2}{*}{Target} & \multirow{2}{*}{Method}
        & \multicolumn{2}{c}{GQA}
        & \multicolumn{2}{c}{F30K}
        & \multicolumn{2}{c}{TextVQA}
        & \multicolumn{2}{c}{CharXiv}
        & \multicolumn{2}{c}{MMMU}
        & \multicolumn{2}{c}{MT}
        & \multicolumn{2}{c}{Overall} \\
        \cmidrule(lr){3-4}\cmidrule(lr){5-6}\cmidrule(lr){7-8}
        \cmidrule(lr){9-10}\cmidrule(lr){11-12}\cmidrule(lr){13-14}
        \cmidrule(lr){15-16}
        & & MAT & Speedup & MAT & Speedup & MAT & Speedup & MAT & Speedup & MAT & Speedup & MAT & Speedup & MAT & Speedup \\
        \midrule
        \multirow{3}{*}{Q3VL-4B}
        & AR  & -- & 1.00$\times$ & -- & 1.00$\times$ & -- & 1.00$\times$
             & -- & 1.00$\times$ & -- & 1.00$\times$ & -- & 1.00$\times$
             & -- & 1.00$\times$ \\
        & EAGLE-3 & \underline{1.95} & \underline{0.71$\times$}
                  & \underline{1.96} & \underline{0.67$\times$}
                  & \underline{1.98} & \underline{0.71$\times$}
                  & \underline{2.28} & \underline{0.76$\times$}
                  & \textbf{2.41} & \underline{0.81$\times$}
                  & \textbf{2.14} & \underline{0.60$\times$}
                  & \underline{2.12} & \underline{0.71$\times$} \\
        & DFlash$^{\dagger}$ & \textbf{3.05} & \textbf{2.28$\times$}
                 & \textbf{2.84} & \textbf{2.36$\times$}
                 & \textbf{2.05} & \textbf{1.57$\times$}
                 & \textbf{2.63} & \textbf{2.06$\times$}
                 & \underline{2.28} & \textbf{1.74$\times$}
                 & \underline{2.13} & \textbf{1.70$\times$}
                 & \textbf{2.50} & \textbf{1.95$\times$} \\
        \midrule
        \multirow{3}{*}{Q3VL-8B}
        & AR & -- & 1.00$\times$ & -- & 1.00$\times$ & -- & 1.00$\times$
             & -- & 1.00$\times$ & -- & 1.00$\times$ & -- & 1.00$\times$
             & -- & 1.00$\times$ \\
        & EAGLE-3 & \underline{2.22} & \underline{0.85$\times$}
                  & \underline{2.20} & \underline{0.85$\times$}
                  & \underline{2.05} & \underline{0.81$\times$}
                  & \underline{2.66} & \underline{0.99$\times$}
                  & \underline{2.48} & \underline{0.95$\times$}
                  & \textbf{2.17} & \underline{0.85$\times$}
                  & \underline{2.30} & \underline{0.88$\times$} \\
        & DFlash$^{\dagger}$& \textbf{2.45} & \textbf{1.78$\times$}
                 & \textbf{2.89} & \textbf{2.18$\times$}
                 & \textbf{2.53} & \textbf{1.87$\times$}
                 & \textbf{3.45} & \textbf{2.73$\times$}
                 & \textbf{3.09} & \textbf{2.55$\times$}
                 & \underline{2.08} & \textbf{1.72$\times$}
                 & \textbf{2.75} & \textbf{2.14$\times$} \\
        \midrule
        \multirow{5}{*}{Q3.6-27B}
        & AR & -- & 1.00$\times$ & -- & 1.00$\times$ & -- & 1.00$\times$
             & -- & 1.00$\times$ & -- & 1.00$\times$ & -- & 1.00$\times$
             & -- & 1.00$\times$ \\
        & EAGLE-3 & 2.37 & 1.74$\times$ & 2.19 & 1.64$\times$ & 2.38 & 1.70$\times$
                  & 2.50 & 1.83$\times$ & 2.23 & 1.64$\times$ & 2.20 & 1.62$\times$
                  & 2.31 & 1.69$\times$ \\
        & MTP & 3.22 & 1.64$\times$ & 3.16 & 1.63$\times$ & 3.26 & 1.64$\times$
              & 3.39 & 1.72$\times$ & 3.32 & 1.71$\times$ & 2.99 & 1.50$\times$
              & 3.22 & 1.64$\times$ \\
        & DFlash & \textbf{4.19} & \textbf{2.42$\times$}
                 & \textbf{4.30} & \textbf{2.12$\times$}
                 & \textbf{4.16} & \textbf{2.53$\times$}
                 & \textbf{5.12} & \textbf{3.19$\times$}
                 & \textbf{4.35} & \textbf{3.01$\times$}
                 & \textbf{4.17} & \textbf{2.30$\times$}
                 & \textbf{4.38} & \textbf{2.60$\times$} \\
        & DSpark& \underline{3.27} & \underline{1.92$\times$}
                 & \underline{3.26} & \underline{1.69$\times$}
                 & \underline{3.36} & \underline{1.98$\times$}
                 & \underline{4.42} & \underline{2.53$\times$}
                 & \underline{3.44} & \underline{2.34$\times$}
                 & \underline{3.35} & \underline{1.76$\times$}
                 & \underline{3.52} & \underline{2.04$\times$} \\
        \bottomrule
    \end{tabular}%
    }
\end{table*}

Table~\ref{tab:main_results} extends the Qwen3.6-27B comparison to the
Qwen3-VL family. On Qwen3.6-27B, under a matched SGLang backend, evaluation
examples, and decoding protocol, DFlash achieves the highest MAT and speedup
on every benchmark, with a task-equal MAT of 4.38 and an average speedup of
$2.60\times$ over AR; DSpark ranks second with 3.52 MAT and $2.04\times$
speedup. On the matched HuggingFace (HF) subsets$^{\dagger}$, DFlash reaches
2.50 MAT and $1.95\times$ speedup for Qwen3-VL-4B, and 2.75 MAT and
$2.14\times$ speedup for Qwen3-VL-8B. By contrast, the SGLang EAGLE-3 runs
remain below their matched SGLang AR baselines, at $0.71\times$ and
$0.88\times$ overall, respectively. Because the Qwen3-VL EAGLE-3 and DFlash
measurements use different backends, their absolute latency and throughput
values are not compared directly; each reported speedup is normalized only to
its own backend-matched AR run.

Four observations stand out.

\textbf{First, drafter capacity appears to raise the attainable acceptance
ceiling}: DFlash uses approximately 1.73B parameters, compared with
approximately 0.43B for Qwen3.6's native MTP module and 0.60B for EAGLE-3,
and produces substantially longer accepted sequences. However, EAGLE-3 does
not outperform the smaller MTP module, showing that scale alone is insufficient
and that drafting architecture and training objective also matter.

\textbf{Second, checkpoint provenance matters.} Despite using a larger drafter,
the third-party DSpark checkpoint trails DFlash on all six multimodal tasks.
This gap contrasts with the strong text-only results reported in
DSpark~\citep{cheng2026dspark}, suggesting that text performance does not
reliably predict multimodal acceptance quality.

\textbf{Third, natively multimodal-trained models better support speculative decoding.}
Qwen3.6-27B, which is natively trained with multimodal data, consistently benefits from speculative decoding across different methods, achieving $1.69\times$ with EAGLE-3, $2.04\times$ with DSpark, and $2.60\times$ with DFlash on the SGLang backend. In contrast, on the Qwen3-VL family, EAGLE-3 fails to provide speedup on SGLang ($0.71\times$ for 4B and $0.88\times$ for 8B). Even when we explicitly train a DFlash-style block-parallel drafter for Qwen3-VL-8B with multimodal data using SpecForge, it reaches only $2.14\times$ speedup on HF, below the $2.60\times$ achieved by the off-the-shelf z-lab DFlash checkpoint on Qwen3.6-27B.
These results suggest that native multimodal training may induce a more unified feature distribution and better alignment between multimodal representations and future-token prediction, making the target model inherently easier for speculative drafters to approximate than adding multimodal signals only during post-hoc drafter training.

\textbf{Fourth, larger target models benefit more from speculative decoding.}
We observe a clear improvement as the target model grows. 
For DFlash, speedup increases from $1.95\times$ on Qwen3-VL-4B to $2.14\times$ on Qwen3-VL-8B, and reaches $2.60\times$ on Qwen3.6-27B. 
EAGLE-3 shows an even sharper transition, from no net acceleration on 4B ($0.71\times$) and 8B ($0.88\times$) to $1.69\times$ on 27B. Two factors likely contribute. 
First, DFlash uses a shallow, roughly fixed-cost 5-layer drafter, while the target-model forward cost grows substantially with model size; the drafting overhead is therefore increasingly amortized by each accepted block. 
Second, larger targets also appear easier to speculate: DFlash MAT rises from $2.50$ and $2.75$ on 4B and 8B to $4.38$ on 27B. 
This suggests that larger models may expose more stable and predictable feature distributions, allowing the drafter to better match future-token representations. Larger targets thus benefit from both a more favorable compute ratio and higher draft predictability.



\subsection{Does Multimodal Conditioning Remain a Bottleneck Under Block Parallelism?}

\begin{figure}[h]
\centering
\includegraphics[width=\linewidth]{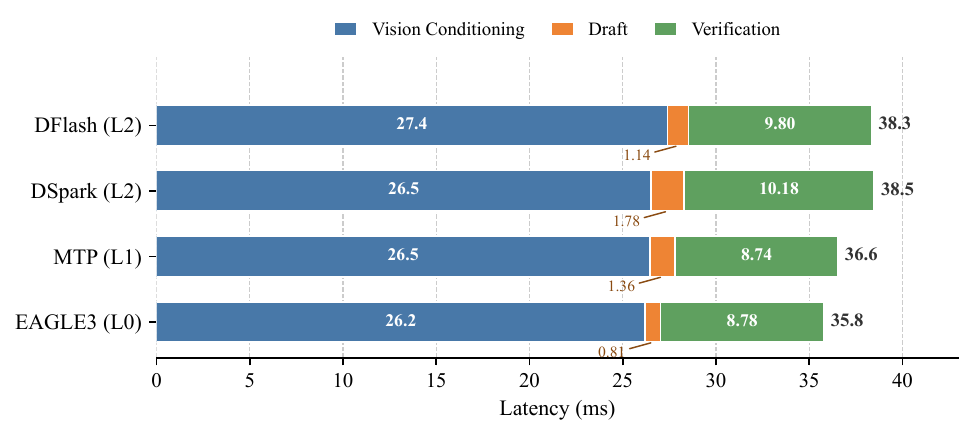}
\caption{Stage-level latency decomposition of a representative decoding pass on Qwen3.6-27B. Vision conditioning includes vision encoding, multimodal prefill, and drafter-side target-feature-to-KV construction when applicable.}
\label{fig:stage_latency}
\end{figure}

\begin{table*}[h]
    \centering
    \caption{HR-Bench results under 4K and 8K resolutions on Qwen3-VL, with a
    512-token generation cap on a common 30-sample manifest. MAT denotes the
    mean number of output tokens advanced per speculative step. Speedup is
    end-to-end latency speedup over the backend-matched autoregressive baseline.
    Top1 denotes top-1 acceptance; Sampling denotes speculate sampling.}
    \label{tab:hr_bench}
    \small
    \setlength{\tabcolsep}{4pt}
    \begin{tabular*}{\textwidth}{@{\extracolsep{\fill}}lcccccccc@{}}
        \toprule
        \multirow{3}{*}{Resolution}
        & \multicolumn{4}{c}{Qwen3-VL-4B}
        & \multicolumn{4}{c}{Qwen3-VL-8B} \\
        \cmidrule(lr){2-5}\cmidrule(l){6-9}
        & \multicolumn{2}{c}{Top1}
        & \multicolumn{2}{c}{Sampling}
        & \multicolumn{2}{c}{Top1}
        & \multicolumn{2}{c}{Sampling} \\
        \cmidrule(lr){2-3}\cmidrule(lr){4-5}
        \cmidrule(lr){6-7}\cmidrule(l){8-9}
        & MAT & Speedup
        & MAT & Speedup
        & MAT & Speedup
        & MAT & Speedup \\
        \midrule
        HR-Bench 4K
        & 2.90 & 1.03$\times$
        & 2.45 & 0.96$\times$
        & 3.22 & 1.06$\times$
        & 2.88 & 1.11$\times$ \\
        HR-Bench 8K
        & 2.86 & 0.90$\times$
        & 2.70 & 0.83$\times$
        & 3.09 & 0.99$\times$
        & 3.22 & 0.85$\times$ \\
        \bottomrule
    \end{tabular*}
\end{table*}

\paragraph{Multimodal prefill, not block drafting, bounds end-to-end speedup.}

A central tension in multimodal speculative decoding is the cost of accessing the visual condition. A condition-light drafter is cheap but poorly aligned with visually grounded continuations; a fully conditioned drafter improves alignment but may duplicate part of the target-side multimodal prefill. Block-parallel drafting changes this trade-off, but does not remove it.

Figure~\ref{fig:stage_latency} makes this clear. For L2 drafters, draft generation itself is no longer the bottleneck: DFlash spends only $1.14$ ms on drafting and DSpark $1.78$ ms. This confirms the main benefit of block parallelism: proposing a longer block is cheap. However, the dominant term is still multimodal conditioning. MTP, DSpark, and DFlash spend $26.45$--$27.40$ ms in the vision-conditioning stage, far larger than the draft step and even larger than target verification. DFlash is especially informative here. Its drafter is fast, but it conditions on target hidden states by projecting them into the drafter KV cache. This target-feature-to-KV construction improves draft quality, yet it also introduces a prefill-like cost that is tied to the visual sequence length. Thus, DFlash removes most of the token-drafting cost, but not the cost of preparing the multimodal condition.

We further stress this effect on HR-Bench, using Qwen3-VL at 4K and 8K resolutions. 
Table~\ref{tab:hr_bench} shows that the drafter still obtains non-trivial MAT, staying around $2.5$--$3.2$ across models and verification rules. 
However, this does not translate into end-to-end acceleration. 
At 4K, speedups are already marginal, ranging from $0.96\times$ to $1.11\times$.
At 8K, all settings fall to or below the autoregressive baseline, with
speedups of $0.90\times$ and $0.83\times$ on Qwen3-VL-4B, and $0.99\times$
and $0.85\times$ on Qwen3-VL-8B. The contrast is especially clear under
rejection sampling on Qwen3-VL-8B: MAT increases from $2.88$ to $3.22$ when
moving from 4K to 8K, yet speedup drops from $1.11\times$ to $0.85\times$.
Thus, the failure mode is not simply poor draft quality. Rather,
higher-resolution inputs increase visual prefill and drafter-side
cache-construction cost, while MAT does not grow enough to amortize this
added conditioning overhead.

These two measurements point to the same conclusion. Under block-parallel speculative decoding, drafting one more token is almost free; conditioning on one more visual token is not. The end-to-end speedup is therefore bounded by multimodal prefill, especially for DFlash-style drafters that additionally convert target features into drafter-side KV cache. Future gains should thus come less from further accelerating the draft forward pass, and more from reducing or reusing the conditioning path: compressed visual tokens, target-feature reuse, shared drafter caches, or lightweight visual-state adapters.

\paragraph{Acceptance is robust to context removal.}

\begin{table*}[t]
    \centering
    \caption{Effect of multimodal target context on DFlash speculative
    acceptance. Results use top-1 verification with 100 examples per task and
    a maximum generation length of 2,048 tokens.  Overall MAT is the equal-weight arithmetic mean over GQA, TextVQA,
    and MMMU. Text-only context masks visual-token K/V entries in the DFlash
    context, whereas no-prefill context masks the complete original prefill
    context. The target model always receives the full multimodal input.}
    \label{tab:dflash_context_ablation}
    \small
    \setlength{\tabcolsep}{4pt}
    \begin{tabular*}{\textwidth}{@{\extracolsep{\fill}}llccccc@{}}
        \toprule
        Model & DFlash context setting & GQA & TextVQA & MMMU
        & Overall MAT $\uparrow$ & $\Delta$ Overall vs.\ full \\
        \midrule
        \multirow{3}{*}{Qwen3-VL-8B}
        & Full multimodal context & 2.692 & 2.750 & 3.259 & \textbf{2.900} & -- \\
        & Text-only context       & 2.608 & 2.504 & 3.246 & 2.786 & $-0.114$ ($-3.9\%$) \\
        & No-prefill context      & 2.845 & 1.953 & 2.946 & 2.581 & $-0.319$ ($-11.0\%$) \\
        \midrule
        \multirow{3}{*}{Qwen3-VL-4B}
        & Full multimodal context & 2.587 & 2.305 & 2.583 & \textbf{2.491} & -- \\
        & Text-only context       & 2.487 & 2.221 & 2.453 & 2.387 & $-0.104$ ($-4.2\%$) \\
        & No-prefill context      & 2.095 & 1.904 & 2.417 & 2.139 & $-0.353$ ($-14.1\%$) \\
        \bottomrule
    \end{tabular*}
\end{table*}

Table~\ref{tab:dflash_context_ablation} further clarifies what kind of
bottleneck multimodal conditioning creates under block parallelism. DFlash
acceptance is only weakly sensitive to explicit visual conditioning in the
drafter context: masking visual-token K/Vs reduces Overall MAT by only
$4.2\%$ on Qwen3-VL-4B and $3.9\%$ on Qwen3-VL-8B. Even removing the entire
original prefill context lowers MAT by only $14.1\%$ and $11.0\%$,
respectively. Since the target model always receives the full multimodal
input, this ablation isolates the drafter-side source of acceptance. The
results suggest that the diffusion-based drafter does not exploit visual
evidence in the same way as a full VLM target during autoregressive
generation. Instead, a large fraction of accepted draft tokens can be
predicted from local continuation statistics and recent contextual states,
rather than from detailed re-access to the full question-image input.

The degradation is also task dependent rather than uniform. TextVQA is the
most sensitive to context removal, especially on Qwen3-VL-8B ($2.750$ to
$1.953$ under no-prefill context), whereas MMMU remains comparatively stable
and GQA shows no consistent monotonic drop. This pattern suggests that
explicit source conditioning matters most when the next-token distribution is
tightly constrained by localized visual or OCR evidence, but contributes less
when acceptance is dominated by short-range target-token continuation. A
larger target also appears slightly more robust to drafter-side context
ablation, which may indicate that stronger models expose more stable local
continuation patterns for speculative matching.

\subsection{Training and Inference Framework Support}

Beyond algorithmic effectiveness, the practical adoption of L2 drafting also
depends on the availability of training and inference infrastructure.
Block-parallel drafting requires dedicated support for drafter training,
block-level proposal generation, and speculative verification. We therefore
summarize the current open-source ecosystem for representative L2 methods,
focusing on both training frameworks and inference backends.

As summarized in Table~\ref{tab:L2_framework_support}, support for L2 drafting
has expanded rapidly beyond method-specific research implementations. The
official DFlash repository provides released checkpoints across multiple model
families, including Qwen3, Qwen3.5/3.6, and Gemma, together with inference
support through Transformers, MLX, SGLang, and vLLM. DeepSpec provides a
unified training and evaluation pipeline for DFlash and DSpark on Qwen3 and
Gemma models. AngelSpec supports several block-parallel architectures,
including DFly, DFlash, DFlare, and DSpark, within a unified training
framework using vLLM, SGLang, or HuggingFace-based inference backends.
Speculators provides DFlash and DSpark training with direct integration into
the vLLM ecosystem, while SpecForge provides corresponding support for
DFlash, Domino, and DSpark with direct integration into SGLang. The model
families listed in the table indicate representative documented configurations
rather than exhaustive model--method combinations.










\begin{table*}[t]
\centering
\fontsize{8.5pt}{10pt}\selectfont
\setlength{\tabcolsep}{4pt}
\renewcommand{\arraystretch}{1.2}

\caption{Representative framework support for L2 speculative decoding.}
\label{tab:L2_framework_support}

\begin{tabularx}{\textwidth}{
    >{\raggedright\arraybackslash}p{0.14\textwidth}
    >{\raggedright\arraybackslash}p{0.23\textwidth}
    >{\raggedright\arraybackslash}p{0.27\textwidth}
    >{\raggedright\arraybackslash}X
}
\hline
\textbf{Repository} &
\textbf{Models} &
\textbf{L2 Methods} &
\textbf{Framework Support} \\
\hline

DFlash &
Qwen3/3.5/3.6, Gemma4 &
DFlash &
MLX, SGLang, vLLM \\

DeepSpec &
Qwen3, Gemma4 &
DFlash, DSpark &
Train, Eval \\

AngelSpec &
Qwen3, Hunyuan3 &
DFly, DFlash, DFlare, DSpark &
Train, vLLM, SGLang \\

Speculators &
Qwen3/3.6, Gemma4 &
DFlash, DSpark &
Train, vLLM \\

SpecForge &
Qwen3/3.6 &
DFlash, Domino, DSpark &
Train, SGLang \\

\hline
\end{tabularx}
\end{table*}

At the inference level, vLLM and SGLang provide the most complete
production-oriented support for block-parallel speculative decoding. Both
frameworks support DFlash-style drafting, and support for newer L2 methods
such as DSpark is also being integrated into their speculative-decoding
stacks. Transformers provides a more general assisted-decoding interface and
has begun to incorporate model-specific DFlash support. MLX provides an
additional execution path for DFlash on Apple Silicon, although its L2
coverage is currently narrower than that of vLLM and SGLang.

Overall, L2 speculative decoding is transitioning from isolated research
implementations toward integrated training--serving ecosystems. In particular,
the Speculators--vLLM and SpecForge--SGLang stacks provide increasingly
complete paths from drafter training to efficient inference. However, support
remains uneven across model families and is still substantially less mature
for multimodal targets than for text-only models. This ecosystem gap
constitutes an additional practical barrier to the broader adoption of
multimodal L2 drafting.

\section{L2 Readiness and Future Directions}
\label{sec:readiness}

\subsection{What Determines L2 Readiness?}

Taken together, the results in Section~4 suggest that L2 readiness should not
be treated as a fixed property of an entire modality or model family. Instead,
it depends on the specific target model, multimodal condition, task and input
regime, and inference environment. We therefore interpret readiness at the
model--task--condition--system level.

\paragraph{Model compatibility.}

L2 drafting is effective across the evaluated targets, but its benefit varies
substantially across model families. DFlash achieves positive end-to-end
speedups on both Qwen3-VL-4B and Qwen3-VL-8B, while obtaining substantially
higher MAT and speedup on Qwen3.6-27B. These results indicate that multimodal
L2 drafting is feasible, but not uniformly effective across targets.
Importantly, the observed differences should not yet be attributed to a
single architectural factor, since model scale, drafter checkpoint,
training recipe, and inference backend also vary across configurations.
The current evidence therefore supports \emph{model-dependent compatibility}
rather than an architecture-agnostic notion of L2 readiness.

\paragraph{Multimodal conditioning.}

The strongest multimodal-specific bottleneck is the cost of conditioning.
The latency decomposition in Figure~\ref{fig:stage_latency} shows that
block-parallel draft generation itself is not the dominant cost in our
evaluated configurations, whereas multimodal conditioning contributes a
substantial portion of the speculative decoding overhead. Our conditioning
analysis further suggests that repeatedly exposing the drafter to the full
multimodal context is not always necessary for maintaining useful speculative
acceptance. A L2-ready multimodal system should therefore provide a lightweight
conditioning path that preserves the information needed for drafting without
reproducing the target model's multimodal processing cost.

\paragraph{Task and input dependence.}

Readiness also varies with the workload. Table~\ref{tab:main_results} shows
noticeable task-level variation in MAT and end-to-end speedup, indicating that
the predictability of future tokens differs across captioning, VQA, OCR, and
reasoning tasks. More importantly, the HR-Bench results show that relatively
high MAT does not guarantee acceleration under expensive visual inputs:
increasing the image resolution from 4K to 8K preserves MAT at roughly the
same level while reducing end-to-end speedup to around or below the
autoregressive baseline. Thus, L2 readiness depends not only on draft quality,
but also on whether the accepted-token gain is sufficient to amortize the
multimodal input cost. Dynamic routing across P levels or block sizes is a
promising consequence of this observation, although we leave generation-stage
routing to future work.

\paragraph{System and ecosystem support.}

Practical readiness further depends on whether L2 methods can be trained and
deployed using existing infrastructure. As summarized in
Table~\ref{tab:L2_framework_support}, DFlash and DSpark are increasingly
supported by dedicated training frameworks such as DeepSpec, AngelSpec,
Speculators, and SpecForge, while vLLM and SGLang provide increasingly mature
serving paths for block-parallel speculative decoding. Transformers and MLX
also provide more limited L2 execution paths. This trend shows that L2 is
moving beyond isolated research implementations toward integrated
training--serving ecosystems. However, support remains substantially less
uniform for multimodal targets than for text-only models.

\paragraph{Overall readiness.}

The evidence therefore supports a qualified conclusion: multimodal
speculative decoding is \emph{partially ready} for L2 block-parallel drafting.
L2 can already provide substantial gains for favorable target models and
workloads, demonstrating that multimodality itself is not a fundamental
barrier. However, these gains remain sensitive to target-model compatibility,
multimodal conditioning cost, input characteristics, and framework support.
The practical question is therefore not whether multimodal L2 works in
general, but under which model--task--condition--system configurations its
accepted-token gains are large enough to translate into end-to-end
acceleration.

\subsection{Challenges and Opportunities}

\paragraph{Lightweight multimodal condition injection.}

Future drafters should learn the smallest multimodal representation sufficient for next-block prediction. Promising routes include target-produced semantic summaries, query-based visual compression, layer-adaptive feature selection, and dynamic routing that enables visual conditioning only when token uncertainty indicates that it is needed. The objective should jointly optimize condition sufficiency and added prefill latency.

\paragraph{Dependency modeling for L2 drafting.}

L2 drafters need enough block-internal structure to preserve consistency without returning to $K$ serial steps. Lightweight causal correction, confidence-ordered refinement, local remasking, and cross-position state transfer may offer a better latency--quality balance than either fully independent or fully autoregressive drafting.

\paragraph{Dynamic Parallel levels and block sizes.}

No single drafter is likely to dominate all multimodal tasks. A practical system can route between an autoregressive drafter, multi-token heads (L1), DFlash (L2), DSpark (L2), retrieval, and tree expansion according to task type, current entropy, generation stage, and measured break-even cost. When the predicted gain falls below zero, the system should shorten the block or revert to target autoregressive decoding.

\paragraph{Tree-aware parallel-drafter training.}

Position-wise imitation does not directly optimize the accepted prefix or the utility of a finite verification tree. In block-parallel speculative decoding methods like DFlash, where the drafter is heavily conditioned on target model representations (via multi-layer hidden state injection), the target's top-1 token is frequently covered within the drafter's top-k candidates. For instance, the top-6 draft tokens can already cover 98–99\% of the target probability mass, making exact position-wise matching largely unnecessary. Consequently, training objectives should instead reward recoverable target-consistent paths, calibrate marginal probabilities for budgeted tree construction, and concentrate capacity on frontier positions that limit acceptance. This bridges multimodal L2 drafting with recent tree-based verification methods, rather than treating the drafter and verifier as independent modules.

\paragraph{Multimodal verification and relaxed acceptance.}

Exact token verification is appropriate for lossless text generation, but actions, codec tokens, and visual codebooks may admit functionally or perceptually equivalent outputs. Future systems should define output-space-specific acceptance rules together with explicit quality guarantees, and should distinguish relaxed verification from merely skipping target computation.

\paragraph{Standardized evaluation and system co-design.}

Parallel drafting changes the workload presented to the target: longer verification sequences and wider trees can improve arithmetic intensity but also increase memory traffic and batch interference. A shared benchmark should span architectures, modalities, input lengths, task uncertainty, output lengths, batch sizes, hardware, and backends. It should report candidate and task quality, stage latency, throughput, memory, energy, and both decode-only and end-to-end speedup. Efficient tree attention, KV-cache reuse, overlapped execution, and scheduling should be evaluated under the same protocol so that readiness becomes a reproducible model--task--system measurement.

\section{CONCLUSION}
\label{sec:conclusion}


This paper asks whether multimodal speculative decoding is ready to move from
L0/L1 drafting to L2 block-parallel generative drafting. Our modality-centered
review shows that VLM, Video-LM, VLA, Speech/Audio, and Visual-AR research has
already developed strong mechanisms for condition access, token compression,
candidate construction, and task-aware verification. However, systematic
evidence for L2 block-parallel drafting across multimodal architectures remains
limited. Our cross-architecture study directly examines this transition and
reveals a more nuanced picture: neither a universal failure nor a universal
success. While direct L2 transfer can be weak on some multimodal architectures,
newer architectures demonstrate substantially stronger draft--target
compatibility. These findings suggest that multimodal L2 readiness is
configuration-dependent, determined by architectural compatibility, condition
accessibility, task predictability, dependency handling, and system-level
amortization. Therefore, multimodal speculative decoding is partially ready for
diffusion-based parallel drafting, but practical deployment still requires
careful co-design of the model architecture, conditioning pathway, and inference
system.

\paragraph{Conflicts of Interest:}The authors declare no conflicts of interest.


\bibliography{references}
\bibliographystyle{iclr2026_conference}

\end{document}